\documentclass[lettersize,journal]{IEEEtran}
\usepackage{amsmath,amsfonts}
\usepackage{algorithmic}
\usepackage{algorithm}
\usepackage{array}
\usepackage[caption=false,font=normalsize,labelfont=sf,textfont=sf]{subfig}
\usepackage{textcomp}
\usepackage{stfloats}
\usepackage{url}
\usepackage{verbatim}
\usepackage{graphicx}
\usepackage{cite}
\usepackage{booktabs} 
\usepackage{xcolor} 

\usepackage{colortbl}
\usepackage{color}
\definecolor{black}{RGB}{101,101,101}
\definecolor{B}{RGB}{0,0,255}
\definecolor{pink}{RGB}{234,208,209}
\definecolor{R}{RGB}{150,84,84}
\usepackage{amsmath}
\usepackage{gensymb}
\usepackage{mathrsfs}

\usepackage[colorlinks,urlcolor=blue,linkcolor=blue,citecolor=blue]{hyperref}

\usepackage{graphicx}

\begin{document}

\title{FADE: A Dataset for Detecting Falling Objects around Buildings in Video}

\author{Zhigang Tu,~\IEEEmembership{Senior Member,~IEEE,}~Zhengbo Zhang,~Zitao Gao,~Chunluan Zhou,~\\ Junsong Yuan,~\IEEEmembership{Fellow,~IEEE,}~Bo Du,~\IEEEmembership{Senior Member,~IEEE}
\thanks{Corresponding authors: Zhengbo Zhang, Zitao Gao (Email: zhengbozhang.1@gmail.com, gaozitao@whu.edu.cn).}
\thanks{Zhigang Tu, Zhengbo Zhang,  and Zitao Gao are with the State Key Laboratory of
Information Engineering in Surveying, Mapping and Remote Sensing, Wuhan
University, Wuhan 430079, China.}
\thanks{Chunluan Zhou is with Ant Group co Ltd, Beijing 100020, China.}
\thanks{Junsong Yuan is with the Computer Science and Engineering Department,
The State University of New York at Buffalo, Buffalo, NY 14260 USA.}
\thanks{Bo Du is with the School of Computer Science, Wuhan University, Wuhan 430072, China.}}


\newcommand{\zb}[1]{\textcolor[rgb]{0.78, 0.2274, 0.333}{\textbf{#1}}}
\maketitle

\begin{abstract}
Objects falling from buildings, a frequently occurring event in daily life, can cause severe injuries to pedestrians due to the high impact force they exert.
Surveillance cameras are often installed around buildings to detect falling objects, but such detection remains challenging due to the small size and fast motion of the objects. Moreover, the field of falling object detection around buildings (FODB) lacks a large-scale dataset for training learning-based detection methods and for standardized evaluation. To address these challenges, we propose a large and diverse video benchmark dataset named FADE.
Specifically, FADE contains 2,611 videos from 25 scenes, featuring 8 falling object categories, 4 weather conditions, and 4 video resolutions. 
Additionally, we develop a novel detection method for FODB that effectively leverages motion information and generates small-sized yet high-quality detection proposals. The efficacy of our method is evaluated on the proposed FADE dataset by comparing it with state-of-the-art approaches in generic object detection, video object detection, and moving object detection.
The dataset and code are publicly available at \url{https://fadedataset.github.io/FADE.github.io/}.
\end{abstract}


\section{Introduction}
\begin{figure}[!t]
\centering
\includegraphics[width=2.5in]{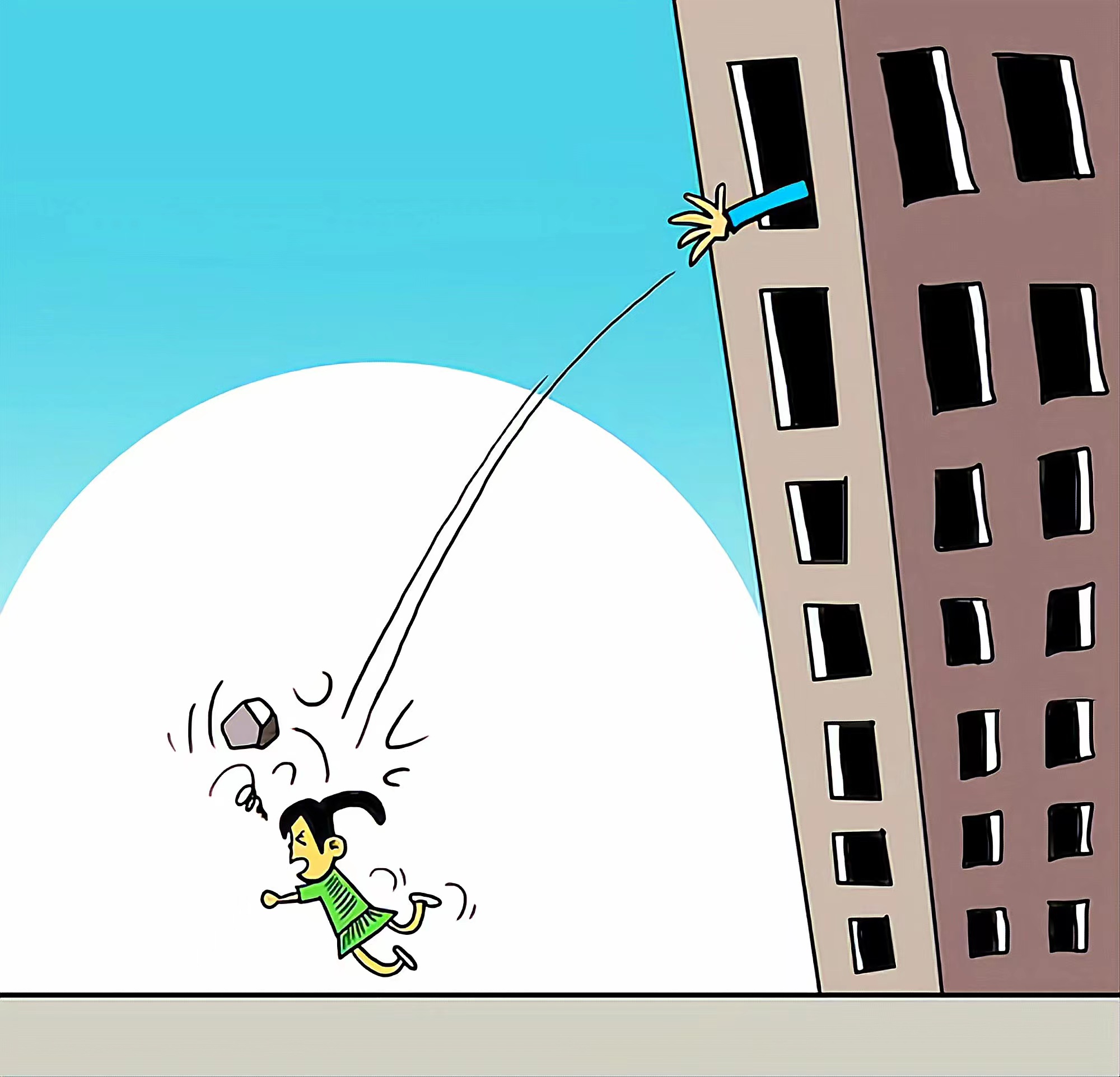}
\caption{Falling object incidents around buildings may pose serious risks to human life and public safety.}
\label{bad}
\end{figure}

\IEEEPARstart{O}{ver} the past few decades, with the continuous expansion of urbanization, high-rise buildings have sprung up. 
Some residents of these buildings throw objects from above without caution, potentially leading to repeated injuries and incidents.
According to a report~\cite{OSHA} of the U.S. Bureau of Labor Statistics, there are more than 50,000 ``struck by falling objects" injuries every year in USA.
To highlight the potential danger posed by objects falling from tall buildings, we present a specific example:
If a 200-gram apple falls from a 30-meter-high building, the impact duration is approximately 0.01 seconds. Based on the momentum theorem~\cite{IC}, this results in an equivalent impact force of roughly 49.5 kilograms. Such incidents can pose serious threats to public safety (see Figure~\ref{bad} for visualization).
To mitigate such incidents, several countries have enacted laws prohibiting the act of throwing objects from buildings, including the USA~\cite{USA}, Singapore~\cite{SG}, and China~\cite{China}.

At the same time, intelligent video surveillance (IVS) has become a critical technology for ensuring public safety~\cite{babaguchi2013guest}, fueled by recent advances in computer vision, including object detection~\cite{aslani2013optical, barnich2010vibe, stauffer1999adaptive}, anomaly detection~\cite{MIL2018, zhou2019anomalynet, chen2020neuroaed, yu2023video}, human identification~\cite{peng2024harnessing,zheng2018pedestrian, navaneet2019operator, du2023video}, tracking~\cite{zhang2024diff,brunetti2018computer, sun2020survey, kumar2020p}, and video understanding~\cite{degardin2021regina, CD-JBF2022, tu2023consistent, zhang2024modular,zhang2025visual}. These computer vision-based IVS methods provide benefits such as low cost, high accuracy, and reduced dependence on manual labor.
Inspired by the successful applications of these techniques and supported by the availability of surveillance cameras around buildings, several research institutions and government agencies have begun exploring IVS algorithms for detecting falling object incidents around buildings (FODB). However, these IVS technologies often rely on large-scale datasets to train learning-based models, yet such a dataset is currently unavailable in the FODB domain. In fact, the FODB task can be regarded as a special case of the moving object detection (MOD) task, and it may appear feasible to utilize existing MOD datasets, such as SABS~\cite{brutzer2011evaluation}, CDnet 2014~\cite{wang2014cdnet}, GTFD~\cite{li2016weighted}, and LASIESTA~\cite{cuevas2016labeled}, to train FODB algorithms. However, this approach is limited by the significant disparity between the two tasks. Specifically, moving (falling) objects in FODB are typically much smaller and move at higher speeds than the objects in standard MOD scenarios.

\begin{figure*}[ht]
  \centering
  \includegraphics[width=1\textwidth]{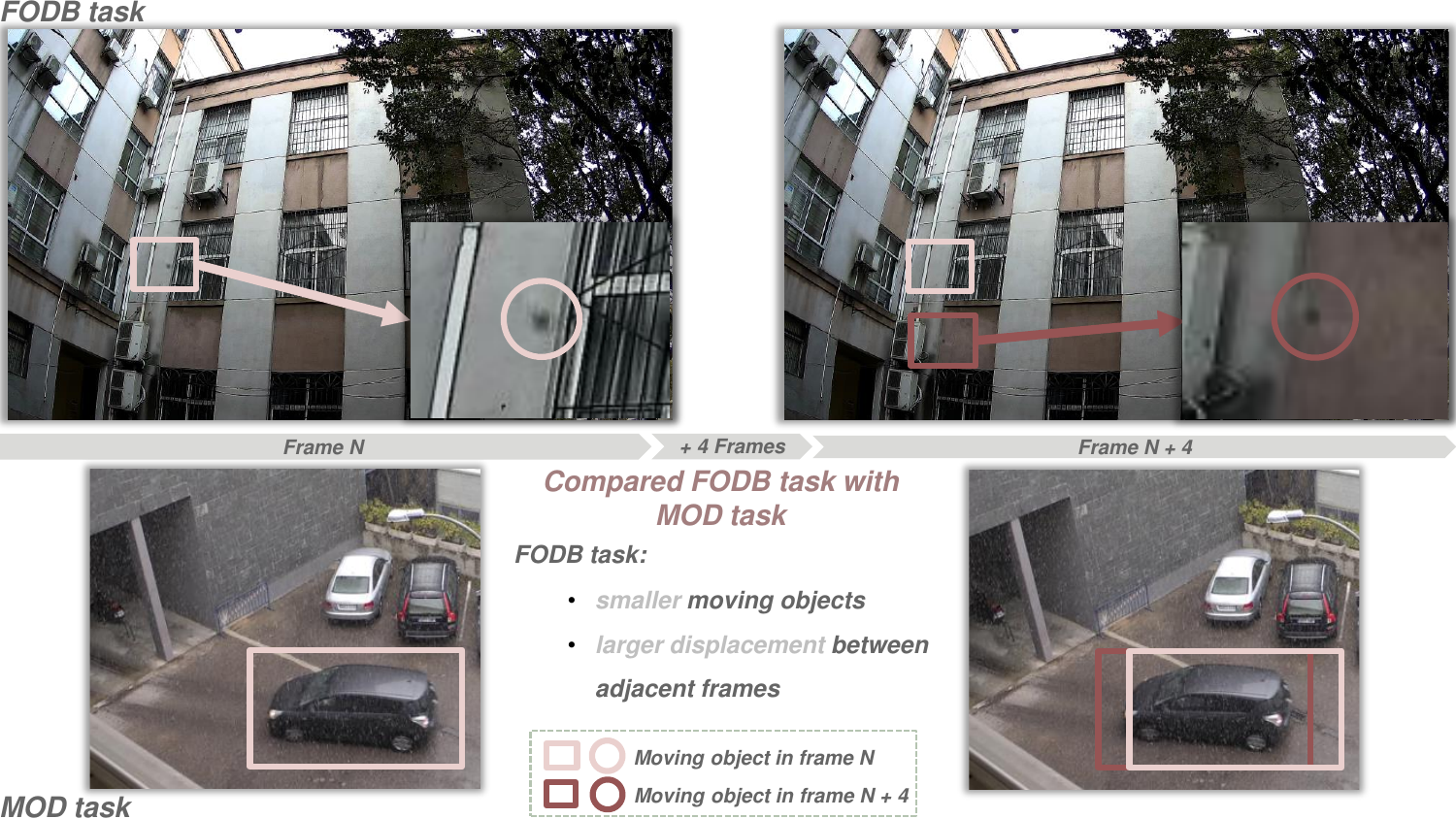}
    \caption{Comparison the tasks of FODB and MOD. By observing four frames (two frames above are from our dataset, and two frames below are from the LASIESTA dataset~\cite{cuevas2016labeled}) from two video sequences, we can find that the moving object in FODB task is much smaller and has larger displacement between succesive frames. To see the falling object in the FODB task clearly, we enlarge the object and place it in the lower right corner of the video frame.}
   \label{compare_annotation}
\end{figure*}

To better illustrate the differences between the FODB task and the MOD task, we present a visual comparison in Figure~\ref{compare_annotation}.
As shown in~Figure~\ref{compare_annotation}, moving objects in existing MOD datasets typically occupy large areas of the image, whereas falling objects in the FODB task are much smaller and cover only a small portion of the scene. Besides, falling objects exhibit rapid motion, resulting in (1) motion blur and (2) large displacements between adjacent video frames. These factors make falling objects difficult to detect and are rarely encountered in MOD datasets.
  Furthermore, none of the existing MOD datasets include categories corresponding to falling objects around buildings. Consequently, it is \textit{necessary} to construct a large and diverse benchmark dataset to evaluate the performance of FODB algorithms. Such efforts are critical for advancing research and facilitating real-world deployment in FODB domain.

\begin{table*}[ht]
\caption{Comparison of our FADE dataset with the prior MOD datasets. ``GT frames" means the Ground Truth frames. 
 The datasets are sorted by the publishing time in ascending order. 
 As can be seen, our FADE dataset has the largest number of videos, total frames, GT frames, and scenes.}
\centering

\begin{tabular}{cccccc}
\toprule
Dataset     & Total videos $\uparrow$ & Total frames & GT frames & Scenes  & Frame-wise IOU (\%)          \\
\midrule
SABS~\cite{brutzer2011evaluation}         & 9            & 6,408        & 2,000     & 9  &    43  \\
SegTrack V2~\cite{li2013video}  & 14           & 976            & 976       & 14    &            \underline{27}  \\
SBI~\cite{maddalena2015towards}          &  14           &  5,024        &  14        &  14    &     31       \\
GTFD~\cite{li2016weighted}          &  25           &  1,067        &  1,067     &  7         &     37    \\
CDnet 2014~\cite{wang2014cdnet}   & 31           & \underline{90,000}        & \underline{90,000}          & 11       &     33     \\
LASIESTA~\cite{cuevas2016labeled}     &  48           &  18,425       &  18,425    &  4      &      35      \\
DAVIS 2016~\cite{perazzi2016benchmark}       & \underline{50}           &  3,455        &  3,455     &  \underline{15}    &   32           \\
\midrule
FADE (Ours)  & \textbf{2,611}          & \textbf{245,177}       &\textbf{245,177}     & \textbf{25} & \textbf{13}         \\ \bottomrule
\end{tabular}
\label{Comparion}
\end{table*}

In this work, we present a video benchmark dataset for FODB, termed FADE, which comprises 2,611 videos captured across diverse scenes. Specifically, the dataset includes 245,177 annotated video frames spanning 25 scenes, 8 object categories, 4 video resolutions, 4 weather conditions, 3 camera angles, and 2 lighting conditions.
As can be seen in Table~\ref{Comparion}, our dataset FADE has richer data compared to the existing MOD datasets. The size of the falling object in our FADE is small  with a median area of about 20 pixels, which is also much smaller than the small object ($area < 32^{2}$ pixels) defined in the COCO dataset~\cite{lin2014microsoft}. We also introduce a new baseline method, FADE-Net, which leverages motion cues to complement appearance features and incorporates a small-object mining region proposal network to generate high-quality proposals for small falling objects. We use the evaluation metrics in the MOD task to benchmark several methods of 3 different tasks (MOD, generic object detection, and video object detection) on our dataset. We also explore a new metric, time range overlap (TRO), to evaluate the performance of the detection methods on localizing the object falling incidents. The experimental results indicate that FODB is a challenging task and validate the effectiveness of our presented method.

The main contributions of our work are three-fold:
\begin{itemize}
\item 
We construct a new video dataset called FADE, which, in terms of application scenarios, is the first dataset for falling object detection around buildings (FODB).
This dataset is large and diverse, which covers various scenes and complex conditions.

\item We explore a new baseline method FADE-Net for the FODB task, which effectively utilizes motion cues and can generate high-quality proposals for small-sized falling objects detection.

\item We evaluate the proposed FADE-Net, and other methods, \textit{i.e.} MOD methods, generic object detection methods and video object detection methods, on our FADE dataset comprehensively, which can be served as a benchmark for future research on FODB.
\end{itemize}

\section{Related Work}

\subsection{Moving Object Detection Dataset}

Moving Object Detection (MOD) is a basic task in computer vision, and some MOD datasets have been released for training and testing the MOD algorithms. Early MOD datasets, such as Wallflower~\cite{1999Wallflower}, SegTrack~\cite{tsai2012motion}, and I2R~\cite{li2004statistical}, are limited in scale. For instance, Wallflower~\cite{1999Wallflower} is one of the earliest MOD datasets, containing only 7 video sequences, each with a single ground truth frame.
 SegTrack~\cite{tsai2012motion} consists of only 6 video sequences and 215 annotated frames.  I2R~\cite{li2004statistical} provides 10 video sequences, including scenes with dynamic backgrounds, challenging weather conditions, and gradual illumination variations.
Later, some large scale and complicated MOD datasets are constructed. For instance, SABS~\cite{brutzer2011evaluation} contains videos with 10 categories, and each video has 800 training frames. Some testing video frames in FBMS 59~\cite{ochs2013segmentation} and SegTrack V2~\cite{li2013video} contain multiple moving objects. GTFD~\cite{li2016weighted} provides a collection of 25 videos with both rigid and non-rigid moving objects. LASIESTA~\cite{cuevas2016labeled} consists of 45 videos and 18,425 video frames, recorded by the moving and static cameras. CDnet 2012~\cite{goyette2012changedetection} and CDnet 2014~\cite{wang2014cdnet} serve as benchmarks for the IEEE Change Detection Workshop, with CDnet 2014 adding 22 videos and 5 new categories to CDnet 2012. Captured in the outdoor environment, BMC 2012~\cite{vacavant2012benchmark} includes a total of 29 real and synthetic videos in different weather conditions. DAVIS 2016~\cite{perazzi2016benchmark} supplies 50 high-quality and densely annotated videos. SBI~\cite{maddalena2015towards} comprises 14 image sequences along with corresponding ground truth backgrounds and is the first dataset designed for evaluating background initialization MOD methods.

Unlike these MOD datasets, FADE is the first dataset for the falling
object detection around buildings task which contains numerous videos and diverse scenes. The quantitative comparison between our FADE and the prior MOD datasets is shown in Table~\ref{Comparion}.

\subsection{Moving Object Detection}
MOD has been extensively investigated due to its wide range of applications~\cite{cucchiara2002sakbot,joshi2012survey,poppe2009moving}. Many unsupervised MOD methods have been explored and can be broadly classified into two categories: background modeling and feature extraction. Specifically, for background modeling, parametric Gaussian mixture methods~\cite{stauffer1999adaptive, varadarajan2013spatial, zivkovic2004improved} have been proposed to represent the background in MOD.
 Barnich~et al. \cite{barnich2010vibe} update the background by applying a novel random selection strategy. Baf~et al.~\cite{el2008fuzzy} apply Choquet integral~\cite{tahani1990information} as an aggregation operator to aggregate color and texture features for MOD.
Lin et al.~\cite{lin2015abandoned} propose a dual-rate background modeling framework for foreground object detection, which leverages both short-term and long-term background models to enhance the accuracy of foreground inference.
 For feature extraction, Yang~et al.~\cite{yang2019unsupervised} adopt an optical flow based method to detect moving objects. Zhou et al.~\cite{zhou2019anomalynet} use motion information to enhance detection with motion fusion blocks that compressing video clips into a single image. Thanikasalam et al.~\cite{thanikasalam2019target} exploit a target-specific Siamese attention network that employs residual and channel attention modules to capture the global and channel-wise information of moving objects. Shang et al.~\cite{shang2017bilinear,vaswani2018robust} consider MOD as a robust principal component analysis problem involving robust subspace learning and tracking. To address sudden and gradual background changes, Dong et al.~\cite{dong2011adaptive} explore a clustering feature space method to represent different background appearances. 

The developments of deep learning~\cite{krizhevsky2012imagenet,long2015fully,ronneberger2015u} have greatly promoted the progress of supervised MOD methods. Early approaches can be broadly classified into six main categories: basic CNN~\cite{braham2016deep,giraldo2020graph,le2018video,patil2020end, zhang2022distilling}, deep CNN~\cite{babaee2018deep,bouwmans2019deep}, 3D CNN~\cite{mandal20193dfr,mandal20203dcd}, ConvLSTM~\cite{chen2017pixelwise}, FCN~\cite{zeng2018background}, and GAN~\cite{sultana2019complete,sultana2019unsupervised}. 
Specifically, CTFU-Net~\cite{xia2024ctfu} hierarchically integrates the local feature extraction capabilities of CNNs with the global context modeling of Transformers to address dynamic scenes. TransBlast~\cite{osman2021transblast} employs an SVD-based subspace loss and Barlow Twins self-supervision to preserve foreground details while reducing the need for extensive annotations.
GraphMOS-U~\cite{prummel2023inductive} and GraphIMOS~\cite{kapoor2025graph} are representative GNN-based approaches. GraphMOS-U~\cite{prummel2023inductive} enables minimal-annotation underwater MOD by initializing Mask R-CNN with domain-specific feature fusion and Sobolev optimization. GraphIMOS~\cite{kapoor2025graph} replaces transductive graphs with inductive block-diagonal GNNs to support real-time deployment on unseen videos.

Although numerous MOD methods have been proposed, few can be applied to the  FODB task. In this work, we provide benchmark results of state-of-the-art MOD methods on our proposed FODB benchmark dataset.

\subsection{Falling Object Detection}

There are a few works~\cite{zheng2014detecting,alvarez2004ultrasonic,yang2022automatic} concentrated on falling object detection. Specifically, \cite{zheng2014detecting} targets the prediction of potential falling objects (much like moving objects) in indoor environments by leveraging 3D point cloud data captured by distance sensors. \cite{alvarez2004ultrasonic} focuses on detecting objects which have already fallen (these are stationary) on railway tracks using ultrasonic sensors and signal coding sequence technology. \cite{yang2022automatic} addresses falling hazards, that is, identifying stationary places a person might fall from or into, by analyzing the positional relationship between hazardous objects and workers at construction sites.

FODB is a critical task, as such incidents occur frequently, with approximately 50,000 cases reported annually in the United States~\cite{OSHA}. These incidents pose fatal risks to pedestrians due to the high impact force of falling objects (see Section~I). However, existing falling object detection methods~\cite{zheng2014detecting, alvarez2004ultrasonic, yang2022automatic} are not designed with small object detection and motion information utilization in mind, making them unsuitable for direct application to FODB scenarios, which involve detecting fast-moving small objects.
 Besides, the FODB field also lacks a public dataset, limiting both the training of learning-based methods and the evaluation of falling object detectors. To solve these issues, we construct the first large-scale benchmark dataset for FODB, and introduce a dedicated FODB method, FADE-Net.

\section{The Proposed FADE Dataset}
\label{sec:dataset}
To advance research in the area of falling object detection around buildings (FODB), we construct and release the first benchmark dataset.
In this section, we provide a detailed introduction to our FADE dataset, covering metadata, dataset construction, dataset splits and statistics, ethical considerations, licensing, maintenance plan, and evaluation metrics.

\begin{figure*}[ht]
  \centering
  \includegraphics[width=1\textwidth]{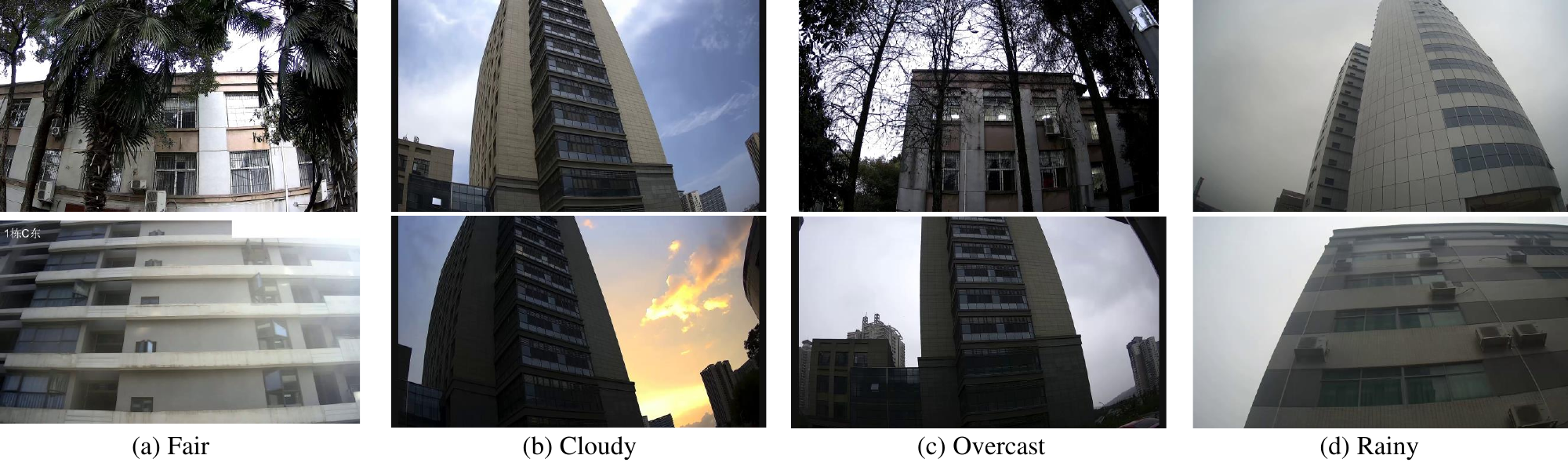}
    \caption{To make our dataset design more accessible, we follow the commonly used Kaggle weather dataset~\cite{narayan2024weather} and categorize weather conditions into four broad types: fair, cloudy, overcast, and rainy.
}
   \label{weather}
\end{figure*}

\subsection{Metadata}
To provide a comprehensive FODB dataset, we collect videos in diverse weather conditions, lighting cases, scenes, camera angles, and video resolutions. Example video frames are shown in our dataset (\url{https://fadedataset.github.io/FADE.github.io/}). Our metadata is defined as follows:

\noindent{\textbf{Object Category.}}
To cover classes of the falling objects around buildings as many as possible, we collect 8 category of objects as follows: clothes, shoes, kitchen waste, books, spitballs, bottles, packaging bags, and packaging boxes.

\noindent{\textbf{Weather Condition}.} Different weather conditions lead to different light intensities, which affect the contrast between falling object and the background. 
To make our dataset design more accessible, we follow the commonly used Kaggle weather dataset~\cite{narayan2024weather} and classify weather conditions into four broad categories: fair, cloudy, overcast, and rainy, as illustrated in Figure~\ref{weather}.

\begin{figure*}[ht]
  \centering
  \includegraphics[width=1\textwidth]{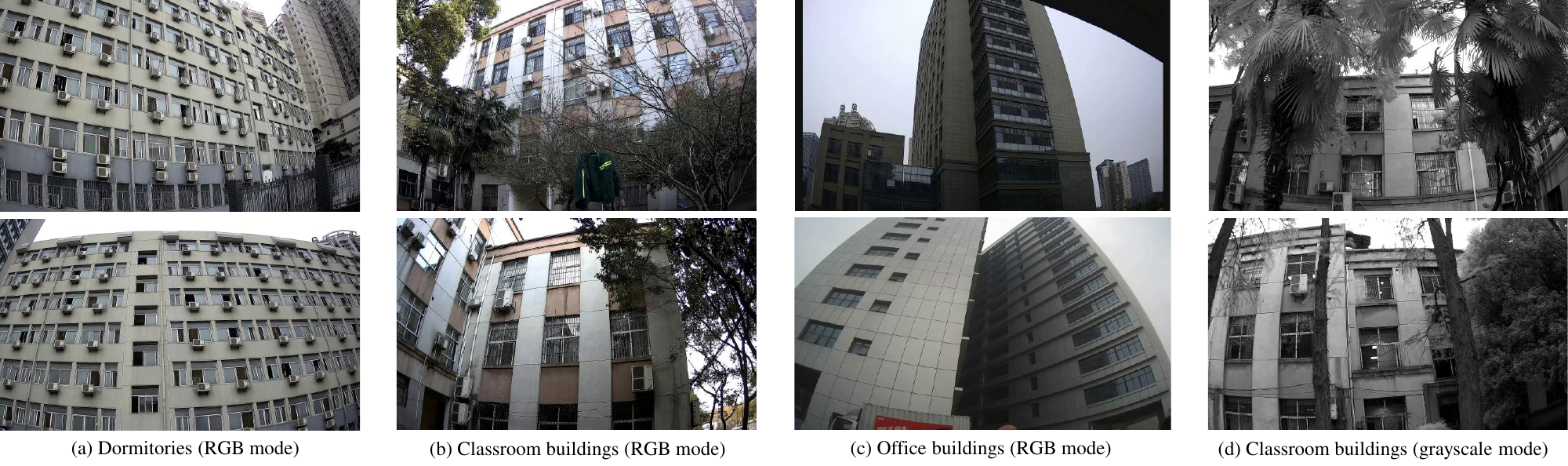}
    \caption{Our dataset includes two modes (RGB and grayscale), captured across various scenes such as dormitories, classroom buildings, and office buildings.
}
   \label{scene}
\end{figure*}

\noindent{\textbf{Lighting Condition.}} Generally, when the light intensity is lower than 0.04 Lux, the image of the surveillance video will change from the RGB mode to the grayscale mode.
Therefore, to ensure the generality of our dataset, we provide videos in both RGB mode and grayscale mode, corresponding to light intensities greater than and less than 0.04 Lux, respectively, as shown in Figure~\ref{scene}.

\noindent \textbf{Scene.} The scenes in our dataset are diverse. Specifically, the FADE dataset includes 18 distinct scenes that cover a wide range of environments where falling incidents may occur, such as classroom buildings, office buildings, dormitories, apartments, and buildings under construction, as illustrated in Figure~\ref{scene}.
Notably, in our dataset, a scene refers to a specific viewpoint of a building. Across different scenes, the buildings appear distinct due to variations in the camera viewpoints.

\noindent{\textbf{Camera Angle.}} 
The videos in our dataset cover 3 different camera angles: 30\degree, 45\degree, and 60\degree. This design reflects real-world FODB surveillance setups, where cameras are typically installed 30 meters away from  buildings, with different floors monitored using cameras positioned at varying angles.

\noindent{\textbf{Video Resolution.}} Usually, the surveillance video contains a variety of resolutions. To include multifarious data, we provide videos of 4 resolutions: $1280 \times 720$, $1920 \times 1080$, $2560 \times 1440$, and $2592 \times 1520$.

\subsection{Dataset construction}
This section provides a detailed description of our dataset construction process, including data preparation, data collection, data format, and data annotation.
 Moreover, we provide dataset documentation, annotation guidelines, intended use cases, structured metadata, example videos, and evaluation code on our website~\url{https://fadedataset.github.io/FADE.github.io/index.html}.

\noindent{\textbf{Data Preparation.}}
The diversity of object categories is important for training models for falling object detection around building. As stated above, the falling objects in our dataset span several common categories, including clothes, shoes, kitchen waste, books, spitballs, bottles, packaging bags, and packaging boxes. In addition, each category contains a diverse set of object instances.
 For example, the kitchen waste category includes items such as banana peels and uneaten apples.

\begin{figure}[!t]
\centering
  \includegraphics[width=0.4\textwidth]{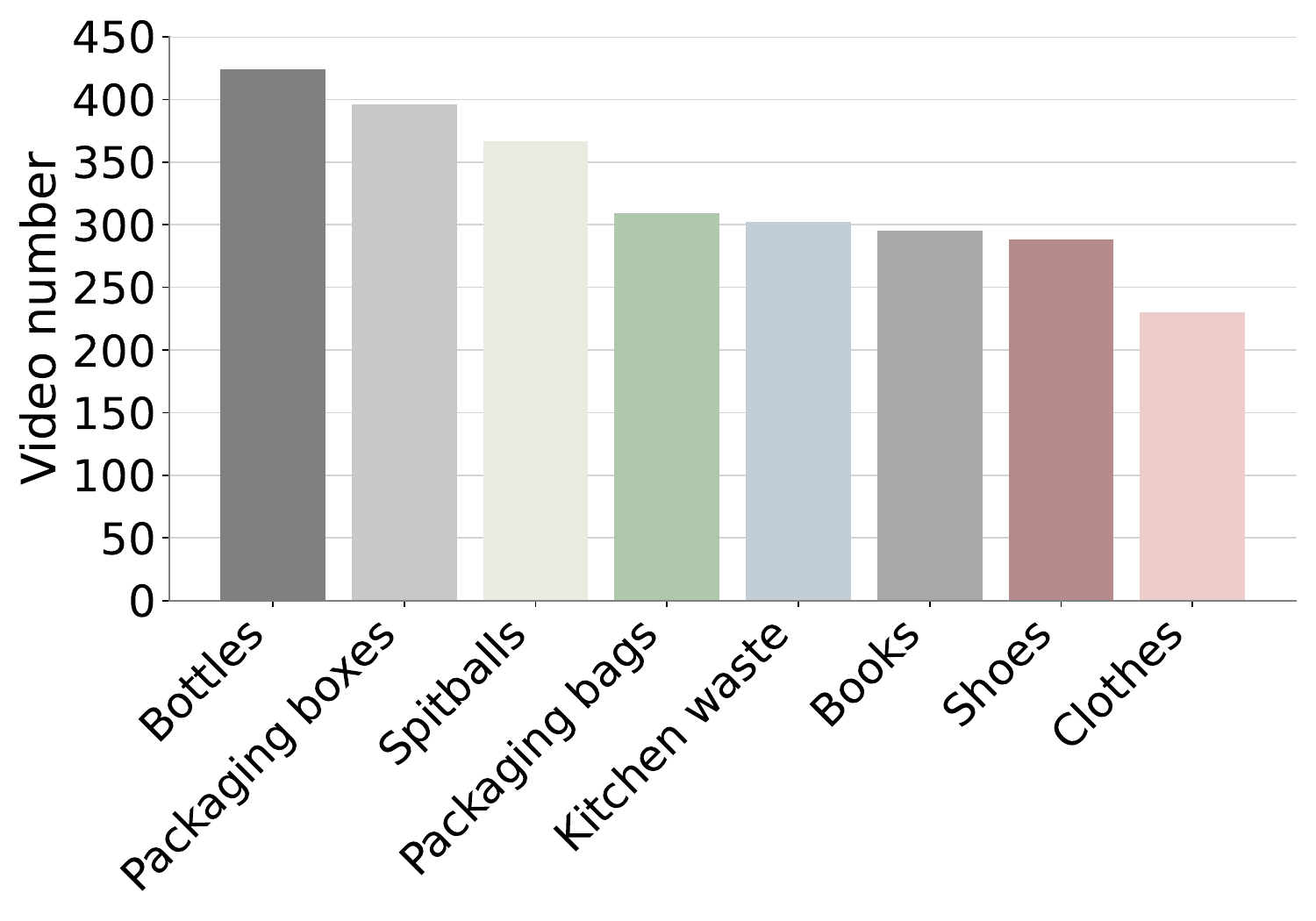}
    \caption{Statistics of each object category's video number in our FADE dataset, sorted by descending order.}
\label{Statistics}
\end{figure}

\noindent{\textbf{Data Collection.}} We recruit seven volunteers and spend a year and a half to collect the video data. Specifically, we throw objects from the high-rise buildings of the corresponding university, and use a wide dynamic range camera, equipped with a $1/3^{''}$ progressive scanning CMOS sensor and dot matrix LED infrared lamp to record the whole process of these events. The highest output video quality of the camera is $2592 \times 1520$ @ $30$ FPS. 
We observe that some falling objects in our recorded video exhibit motion blur due to their high falling speed. Although recording videos with CMOS cameras equipped with global shutters and high FPS can mitigate this issue, the use of such specialized cameras presents two main drawbacks: (1) videos captured with these cameras often suffer from KTC noise~\cite{lauxtermann2007comparison}, introducing visual artifacts; and (2) most existing surveillance systems deployed worldwide for monitoring falling objects use general-purpose, low-cost sensors. As a result, models trained on data from these specialized CMOS cameras may be less applicable to real-world FODB scenarios.

\noindent{\textbf{Data format.}}
The annotation format of our dataset is PASCAL VOC~\cite{everingham2010pascal} style, which is one of the most popular dataset annotation formats. The detailed data format is provided at~\url{https://fadedataset.github.io/FADE.github.io/document.html}.

\noindent{\textbf{Data Annotation.}} The annotation process of our dataset lasts for half a year, including two rounds. In the first round, the novice annotator labels the video. In the second round, the expert annotator checks the annotation to improve the quality. Different from the pixel-by-pixel annotation method of the MOD datasets~\cite{brutzer2011evaluation,li2016weighted,maddalena2015towards}, we adopt the annotation manner used in the object detection task to generate the bounding box around each falling object.

\begin{figure}[!t]
\centering
  \includegraphics[width=0.35\textwidth]{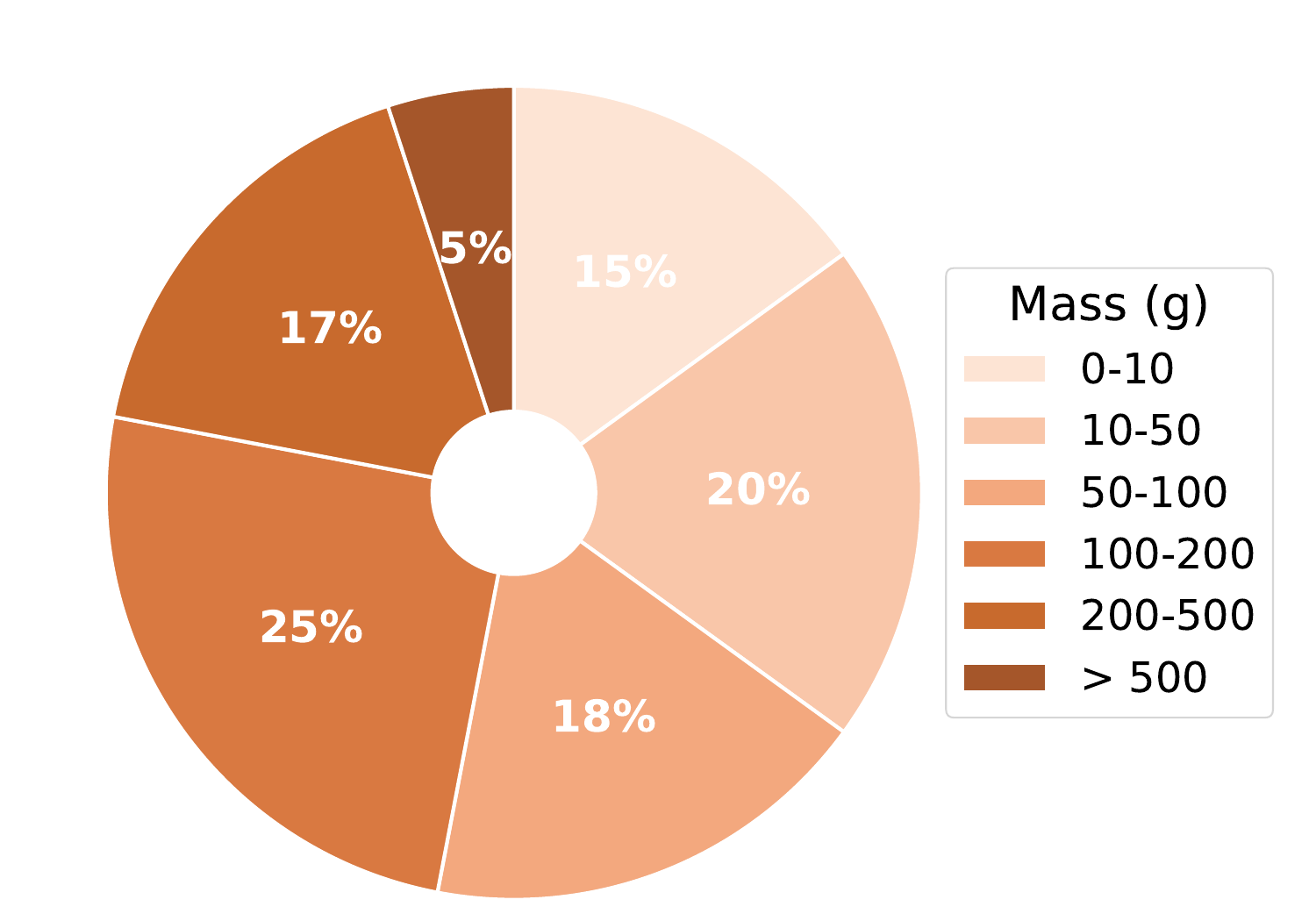}
    \caption{Percentage of falling objects with different mass in our FADE dataset.}
\label{mass}
\end{figure}

\subsection{Dataset division and statistics}
\label{sec:statistics}
To provide a more intuitive understanding of our dataset, we present several quantitative statistics, including the dataset division, falling object sizes, the number of instances per object category, and the proportion of object area relative to the image.

\noindent{\textbf{Dataset Division.}} When splitting the dataset, we consider five attributes for each video: video resolution, scene, lighting condition, weather condition, and object category.
It is worth noting that the videos in our training, validation, and testing sets collectively cover all labels from the five attributes mentioned above.
To better test the generalization  of FODB algorithms, the scenes in our training set, validation set, and testing set (except for the scenes captured in rainy days) are non-overlapped.

\noindent{\textbf{Dataset Statistics.}} We provide some statistical information about the constructed dataset. As shown in Figure~\ref{Statistics}, to make our FADE dataset with various falling object categories, we collect the videos covering falling objects of eight categories. The percentage of falling objects with different weights is shown in Figure~\ref{mass}.

\begin{figure*}[!h]
  \centering
  \includegraphics[width=\textwidth]{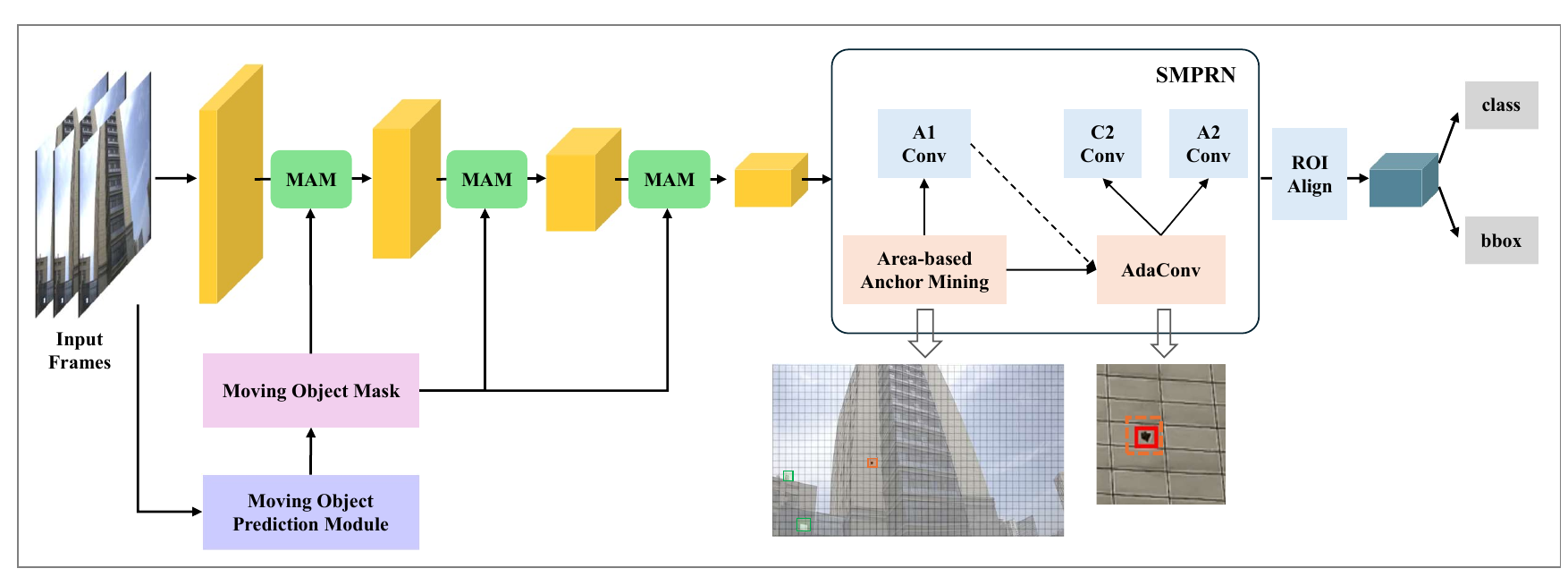}
    \caption{Overview of our proposed method, which is built upon the Faster R-CNN~\cite{ren2015faster} framework with Feature Pyramid Network~\cite{lin2017feature}.} ``MAM'' denotes the Moving attention module. “C” and “A” denote classifier and anchor regressor, respectively. “Conv” and “AdaConv” indicate conventional convolution and the adaptive convolution~\cite{vu2019cascade} layers, respectively.
   \label{pipline}
\end{figure*}

\begin{figure*}[!h]
  \centering
  \includegraphics[width=0.9\textwidth]{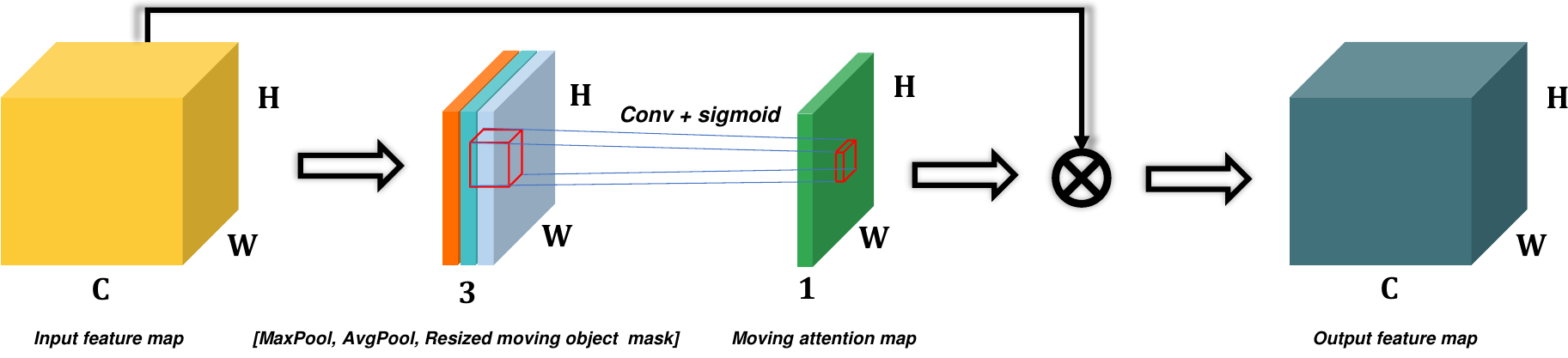}
    \caption{Details of the Moving Attention Module. The Moving Attention Module is designed to fuse the appearance and motion information of objects. To improve the robustness of the appearance features, we apply average pooling and max pooling to the input appearance features before the fusion step.}
   \label{module}
\end{figure*}

\subsection{Dataset ethics, license, and maintenance plan.}

To facilitate the use of our dataset, we provide details regarding its ethical considerations, licensing terms, and maintenance plan in this section.

\noindent{\textbf{Ethics.}} We have full ownership of these videos supplied in our FADE dataset, and all of the collected videos have been authorized for free use by the corresponding university. We inform the seven volunteers, who help us drop the objects from buildings, in advance that their personal information (e.g. faces and hands) may appear in the video, and we get their permission. We conduct two improvements to solve the ethical concerns in our dataset. 1) 
We have reviewed all the videos and found that 32 of them contain people. Among these, 21 videos include appearances of volunteers who assisted in the data collection process.
For the 21 videos, the volunteers are informed of their appearance and have signed on an informed consent form to allow us to use the videos. The link of this signed informed consent form is: \url{https://fadedataset.github.io/FADE.github.io/ethic.html}. For the other 11 videos, we remove them from our FADE dataset. 2) We have obtained authorization from the corresponding university where the video data was captured. All videos in the updated FADE dataset are freely available for research on falling object detection.
We upload the corresponding authorization files to our dataset website (\url{https://fadedataset.github.io/FADE.github.io/ethic.html}). 
Notably, the video in our dataset does not contain any personally identifiable information or offensive content, and we assume full responsibility in the event of any  issues related to data licensing.

\noindent{\textbf{License.}}
Our FADE dataset is published under the CC BY-NC-SA 4.0 license, which means everyone can use our dataset for the non-commercial research purpose. Our code is released under the Apache 2.0 license.

\subsection{Evaluation metrics}
\label{sec:metric}
We use precision, recall, and F-measure to evaluate and compare the performance of methods across different tasks, including MOD, generic object detection, and video object detection.
Since the object size in our dataset FADE is small, we treat a detection as a true positive if its IoU with the GT is larger than 0.3. The F-measure is defined as:
\begin{equation}
F\!-\!measure=\frac{(1+\beta^2)\cdot precision\cdot recall}{\beta^2 \cdot precision + recall}.
\end{equation}

We set the positive real factor $\beta = 1$, when we calculate the F-measure.

Locating the falling incident temporally is crucial to find out the perpetrator who throws the falling object. We therefore design a useful metric named time range overlap (TRO) to evaluate the ability of the algorithm in this regard. The design of TRO is inspired by the DER (Diarization Error Rate)~\cite{yu2022summary} in the speaker diarization task. The TRO is defined as:
\begin{gather}
TRO=\frac{TR_{p} \cap TR_{g t}}{TR_{p} \cup TR_{g t}}, \\ TR_{p} = \lbrack T_{p}^B, T_{p}^E \rbrack, TR_{gt} = [T_{gt}^B, T_{gt}^E],
\end{gather}
where $TR_{p}$ indicates the predicted time range and $TR_{g t}$ denotes the GT time range. $T_{p}^B$ and $T_{gt}^B$ are the predicted and the GT beginning time of a falling incident, respectively. $T_{p}^E$ and $T_{gt}^E$ separately indicates the predicted and the GT ending time of a falling incident.

\section{Proposed Method}
Generic object detection methods rely heavily on appearance features to detect objects. However, in the FODB task, falling objects are typically small and often exhibit blurred appearances due to motion blur caused by high-speed movement, making them difficult to detect using appearance features alone.
 To address this limitation, we propose a method called FADE-Net, built upon Faster R-CNN~\cite{ren2015faster} with FPN~\cite{lin2017feature} (see Figure~\ref{pipline}). Specifically, FADE-Net introduces a Moving Attention Module that incorporates motion cues to complement appearance features for improved detection of fast-moving objects. Additionally, a Small-Object Mining Region Proposal Network (SMRPN) is integrated in our FADE-Net to enhance the detection of small objects common in FODB scenarios. The details of MAM and SMRPN are described in the following sections.

\subsection{Moving attention module (MAM)}
Detecting falling objects around buildings based solely on appearance features is challenging, as these objects are often small and affected by motion blur due to their high speed. To enhance detection accuracy, we leverage the inherent motion characteristics of falling objects by designing a Moving Attention Module that incorporates motion information as a complementary cue to appearance features. Specifically, we first employ a Moving Object Prediction Module to extract a motion mask representing moving objects. This mask is then fed into the proposed Moving Attention Module, where it is fused with appearance features to produce more informative representations for falling object detection. The designs of the Moving Object Prediction Module and the Moving Attention Module are detailed in the following sections.

\noindent \textbf{Moving Object Prediction Module.} To obtain motion information as a complementary cue to appearance features for improved falling object detection, we design a Moving Object Prediction Module. Specifically, to balance robustness and real-time performance, we adopt the MOD method MOG2~\cite{zivkovic2004improved} to generate moving object masks as motion cues. The masks are candidate moving object regions within the current frame.

\noindent \textbf{Moving Attention Module.} The generated moving object mask, which serves as motion information, is fed into the Moving Attention Module and fused with appearance features to provide more robust representations for falling object detection. Specifically, in our Moving Attention Module, the appearance feature is represented by the feature map $F$ from the previous layer, with dimensions $H \times W \times C$. To enhance the robustness of the appearance information, we apply both average pooling and max pooling to $F$ before fusion.

We then concatenate the averaged and max-pooled feature maps with the moving object mask and feed them into a convolutional layer followed by a Sigmoid activation to fuse motion and appearance features. The fused output, after passing through the convolutional layer and Sigmoid activation, forms a more robust representation, referred to as the moving attention map, which serves as the output of our Moving Attention Module. The process is defined as:
\begin{equation}
M = \sigma(\text{Conv}(\text{Concat}(\text{AvgPool}(F), \text{MaxPool}(F), \text{Mask}))),
\end{equation}
where $M$ is the resulting moving attention map, and $\sigma$ denotes the Sigmoid function. It is worth noting that our Moving Attention Module is applied at three locations, each positioned after the downsampling stages in our backbone. All Moving Attention Modules share the same moving object mask.

Overall, the proposed Moving Attention Module enables our detection model to effectively integrate appearance and motion information, resulting in more robust representations for detecting falling objects around buildings.

\subsection{Small-Object Mining RPN (SMRPN)}
After introducing the Moving Attention Module, we now present the Small-Object Mining RPN. While the Moving Attention Module enables the model to leverage both appearance and motion information for more robust detection of small objects, the small size of falling objects may still result in their features being lost during downsampling in the backbone. To address this challenge, we design the Small-Object Mining RPN (SMRPN), which integrates multi-level features and employs dynamic thresholds to ensure that small falling objects can be effectively captured by the model.

Specifically, SMRPN leverages features from all downsampled levels to perform the initial regression, enabling the model to effectively capture information across different object scales. We also adopt an adaptive convolution~\cite{vu2019cascade} to align anchor features with target features, allowing for more refined regression and ultimately generating high-quality proposals for small objects.
In addition, to enhance the model’s ability to detect typically small falling objects, we introduce an area-based anchor mining strategy with a dynamic threshold that adapts to object size in the first stage:
\begin{equation}
\begin{split}
    Threshold=\mathrm {max} \left ( 0.20, 0.15+\alpha \cdot \log_{}{\frac{\sqrt{w\cdot h}}{5}}  \right ), 
 \end{split}
\end{equation}
where $\alpha$ is a scale factor, set to 0.2 by default in our task.

\begin{table*}[ht]
\caption{Performance of different methods on the testing set of our FADE dataset. The best and second best performances are highlighted in \textbf{bold} and \underline{underline}, respectively. DLA34 (DCNv2~\cite{zhu2019deformable})~\cite{yu2018deep} indicates that some of the convolutions in the DLA34 network are replaced by Deformable Convolution v2 (DCNv2). FGFA (FlowNet~\cite{dosovitskiy2015flownet})~\cite{zhu2017flow}, FGFA (PWC-Net~\cite{sun2018pwc})~\cite{zhu2017flow}, and FGFA (RAFT~\cite{teed2020raft})~\cite{zhu2017flow} respectively indicate FGFA based on the optical flow methods of FlowNet, PWC-Net, and RAFT.}
\center
\begin{tabular}{ccccccc}
\toprule
Method    & Type &  F-measure&  Precision &  Recall &  TRO &  FPS  \\ \midrule
FADE-Net (Ours) & MOD-based Generic Object Detection & \textbf{72.08}& \textbf{73.52}& \textbf{70.69}& \textbf{51.77} & 15.7\\ \midrule
Faster R-CNN~\cite{ren2015faster} + FPN~\cite{lin2017feature}& Generic Object Detection &  35.56& 54.55& 26.38& 32.47 & 16.7\\ 
YOLOv5~\cite{YOLOv5} & Generic Object Detection & 33.67& 54.77& 24.31& 34.12 & 32.8\\ 
DLA34 (DCNv2~\cite{zhu2019deformable})~\cite{yu2018deep} & Generic Object Detection & 22.57& 20.04& 25.83&  24.69 & 22.7\\
DETR~\cite{carion2020end}  & Generic Object Detection &  29.98& 49.47& 21.51& 26.80 & 7.9\\
swin-B~\cite{liu2021swin}  & Generic Object Detection &  36.99& 57.37& 27.29& 36.63 & 8.3\\
RT-DETR~\cite{zhao2024detrs} & Generic Object Detection     & \underline{40.15}& \underline{59.24}& 30.36& 38.62 & 28.7\\
\midrule
MOG~\cite{kaewtrakulpong2002improved} & Moving Object Detection & 17.96& 12.85&    29.80&  20.91 & \underline{370.1} \\ 
MOG2~\cite{zivkovic2004improved} & Moving Object Detection & 24.55&  19.03&  34.57&  48.03 & 331.8\\ 
GMG~\cite{godbehere2012visual} & Moving Object Detection &    2.71&   1.55&   10.89&  14.71 & 97.7\\ 
Vibe~\cite{barnich2010vibe} & Moving Object Detection &  15.26&  14.66&  15.91&  23.85 & \textbf{488.5}\\ 
CNT~\cite{CNT} & Moving Object Detection    &      12.84&   19.09&  9.67&   17.25 & 147.3\\ 
FMOD~\cite{rozumnyi2021fmodetect} & Moving Object Detection &  2.48&  6.73&  1.52& 16.09 & 289.7\\ 
KNN~\cite{zivkovic2006efficient} & Moving Object Detection &    0.83&  0.42& 30.48&   34.20 & 188.5\\ 
GSOC~\cite{GSOC}    & Moving Object Detection  &  0.73&   0.38&  10.12&  16.17 & 146.6\\ 
LSBP~\cite{guo2016background}    & Moving Object Detection &     0.20&  0.10&  6.73&  12.35 & 158.0\\ \midrule
MEGA~\cite{chen2020memory} & Video Object Detection & 5.20& 2.71& \underline{65.60}& \underline{48.23} & 8.3\\ 
FGFA~\cite{zhu2017flow} (FlowNet~\cite{dosovitskiy2015flownet}) & Video Object Detection &  0.26& 0.13& 34.01& 46.72 & 6.7\\ 
FGFA~\cite{zhu2017flow} (PWC-Net~\cite{sun2018pwc}) & Video Object Detection &  0.22& 0.11& 31.95& 45.47 & 10.1\\ 
FGFA~\cite{zhu2017flow} (RAFT~\cite{teed2020raft}) & Video Object Detection &  0.18& 0.09& 30.05& 42.97 & 8.8\\ \bottomrule
\end{tabular}

\label{scene and resolution}
\end{table*}

\section{Experiments}
\label{sec:experiment}

In this section, we sequentially present the implementation details of our method, the main results, the ablation study, an analysis of the performance of optical flow-based methods, and visualization results.

\subsection{Implementation details}
Our method is fine-tuned on a pre-trained Faster R-CNN~\cite{ren2015faster} with a ResNet-50~\cite{he2016deep} backbone. We use SGD as the optimizer with a learning rate of 0.005, momentum of 0.9, and weight decay of 0.0005. The model is trained for 15 epochs with a batch size of 2 on our dataset. The Moving Attention Module consists of a convolution layer with a 7 $\times$ 7 kernel followed by a Sigmoid function. Its input includes a moving object mask predicted by MOG2~\cite{zivkovic2004improved} along with its corresponding average pooling map and max pooling map. The kernel size of the convolution layer in our moving attention module is set to $7 \times 7$.

\subsection{Main results}
\label{sec:testing set}
To provide a comprehensive benchmark, we conduct extensive experiments on FADE, evaluating a range of state-of-the-art methods, including our proposed FADE-Net, five generic object detection methods, nine MOD methods, and two video object detection methods. We use the metrics defined in section~\ref{sec:metric} for performance evaluation. We specify the implementation details of all algorithms (requirements, hyperparameters, and training details) at~\url{https://github.com/Zhengbo-Zhang/FADE}.

As can be seen in Table~\ref{scene and resolution}, our proposed FADE-Net achieves the highest performance across all four metrics. During the training process, the CNN generates feature maps that are significantly smaller than the original image, making it is challenging for generic object detection methods to capture features of small objects. 
However, the falling objects in our dataset are of small size. To address this challenge, our FADE-Net leverages multi-stage proposal refinement and an area-based anchor mining strategy, enabling more effective detection of small objects. Furthermore, motion blur caused by the fast motion of falling objects can reduce the recall of image-based detection models. 
To solve it, our FADE-Net integrates the proposed Moving Attention Module, which fuses motion information with appearance features to produce more robust representations for falling object detection.

Although MOG2~\cite{zivkovic2004improved} is the best-performing MOD method, it still shows a substantial precision gap compared to our proposed FADE-Net. This is because FADE-Net effectively extracts and utilizes both appearance and motion features during inference, which significantly enhances falling object detection. In contrast, the MOD method can only update the model online during inference and lacks the ability to exploit various informative features.
This leads MOG2 to detect certain moving objects that are not falling objects, such as clouds and shadows. The detection results of LSBP~\cite{guo2016background} contain numerous ghost regions, such as trailing areas behind fast-moving objects that lie outside their actual contours~\cite{shaikh2014moving}, resulting in the lowest precision and F-measure. MEGA~\cite{chen2020memory} combines global semantic information with local localization cues and leverages more key frames in the video for detection, achieving the second-best recall and TRO performance. However, since MEGA is not designed for small object detection, its network struggles to capture the appearance features of small falling objects. As a result, MEGA yields low precision on the constructed FADE dataset.

The experimental results show that although our method is not the fastest in terms of inference speed, it achieves superior performance owing to the SMRPN and MAM modules, which are specifically designed for FODB. In terms of
performance, our method significantly outperforms the second-best approach (F-measure 72.08 vs 40.15).

\begin{table*}[ht]
\caption{Performance of different modules in the proposed FADE-Net on the testing set of our FADE dataset. We use a GTX 1080 to evaluate the FPS of the algorithm.  The best performances are highlighted in bold.}
\center
\begin{tabular}{cccccccc}
\toprule
Faster R-CNN+FPN& SMRPN& MAM &  F-measure &  Precision &  Recall &  TRO &  FPS  \\ \midrule
\(\surd\) &  &  & 35.62& 54.78& 26.39& 32.50 & \textbf{16.5}\\
\(\surd\) & \(\surd\) & & 53.42& 62.75& 46.51& 39.19 & 16.2\\ 
\(\surd\) &  & \(\surd\) & 61.72& 59.21& 64.46&  47.25 & 15.9\\
\(\surd\) & \(\surd\) & \(\surd\) & \textbf{72.08}& \textbf{73.52}& \textbf{70.69}&  \textbf{51.77} & 15.7\\ \bottomrule
\end{tabular}
\label{ablation}
\end{table*}

\begin{figure*}[ht]
  \centering
  \includegraphics[width=0.95\textwidth]{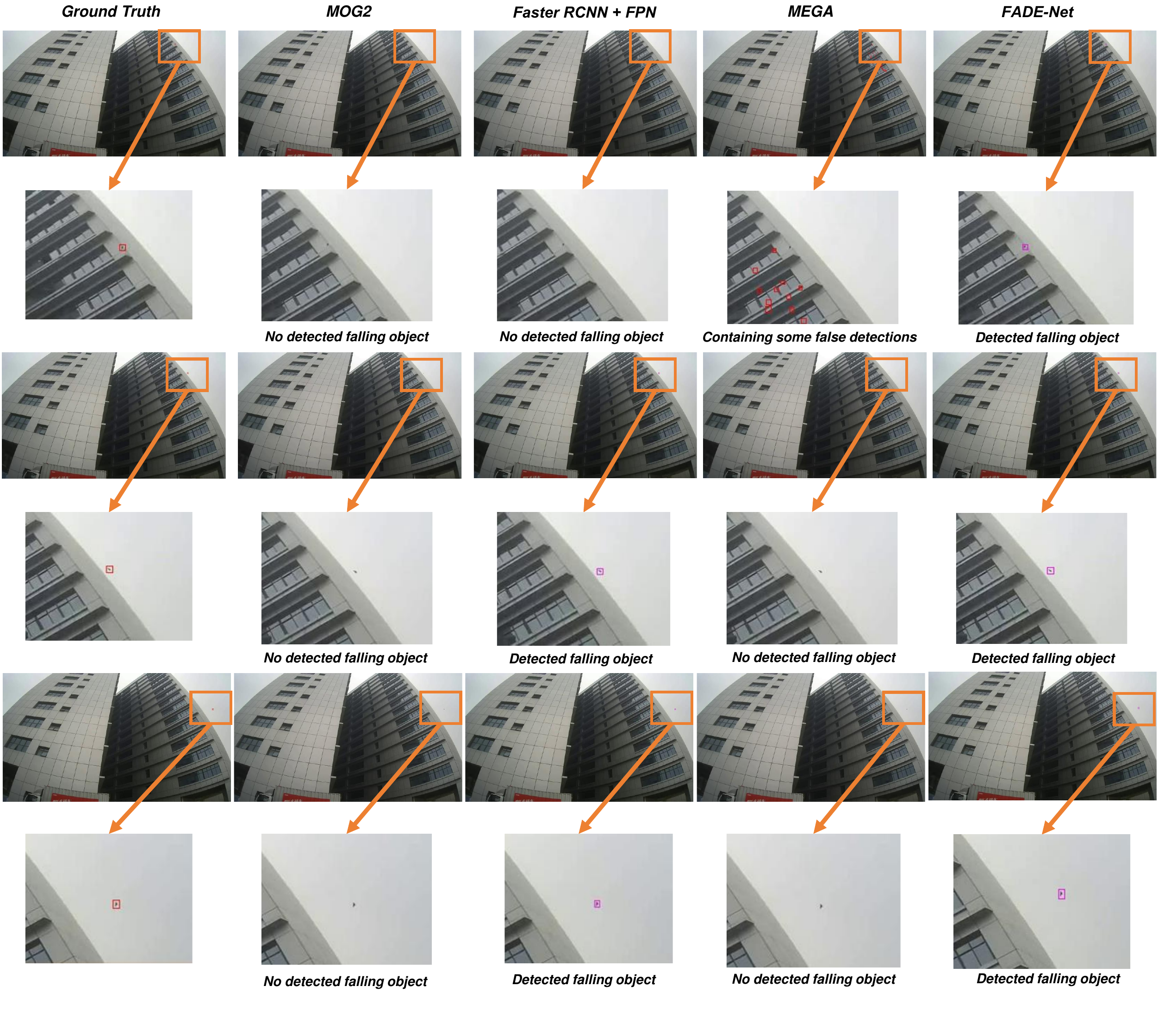}
  \vspace{-8mm}
    \caption{We show ground-truth and the detection results of MOG2~\cite{zivkovic2004improved}, Faster R-CNN~\cite{ren2015faster} + FPN\cite{lin2017feature}, MEGA\cite{chen2020memory} and our proposed FADE-Net on three frames sampled from a video. MOG2, Faster R-CNN + FPN and MEGA are representative algorithms of moving object detection, image object detection, and video object detection respectively. In order to illustrate the difference between the detection results more clearly, we enlarge the regions around the falling objects in the frames.}
      \vspace{-2mm}
   \label{vis}
\end{figure*}

\subsection{Ablation study}
In this part, we conduct ablation studies to evaluate the effectiveness of each module in our proposed method FADE-Net.
As shown in Table~\ref{ablation}, both accuracy and recall are improved when adopting SMRPN. Notably, the recall is boosted by more than 20\% (26.39\% vs 46.51\%).
This indicates that multi-stage cascade refinement enhances the detection accuracy. Additionally, employing a dynamic area-based anchor mining strategy in the initial stage helps avoiding missed detection of small objects.
The adoption of MAM resulted in a significant increase of 38.07\% in the recall rate. 
This indicates that incorporating motion information in our method enables effective capture of high-speed falling objects.
The simultaneous usage of SMRPN and MAM further enhances the detection performance, suggesting that these two modules are complementary. Such improvement also shows that the ability to detect small objects and capture trajectories of moving objects contributes to the effectiveness of falling object detection around buildings. Besides, the experimental results show that introducing the SMRPN and MAM modules has minimal impact on the inference speed of our method. This is because the MAM module is built upon an efficient GPU-accelerated MOG algorithm, while the SMRPN functions primarily as a training-time strategy that employs dynamic thresholds to generate proposals tailored for small object detection.

\subsection{Analysis of optical flow based methods in FODB}
\label{sec:discussion}
Optical flow is good at estimating motion information in video by capturing the pixel-level dense motion field. It is widely used in many video tasks, e.g. action recognition~\cite{carreira2017quo,tu2019action,wang2013action,wang2016temporal} and MOD~\cite{chen2020memory}, and makes good performance. However, as shown in Table~\ref{scene and resolution}, the optical flow based methods FGFA (FlowNet~\cite{dosovitskiy2015flownet})~\cite{zhu2017flow} (the original method using FlowNet to compute optical flow), FGFA (PWC-Net~\cite{sun2018pwc})~\cite{zhu2017flow}, and FGFA (RAFT~\cite{teed2020raft})~\cite{zhu2017flow},  do not obtain the expected performance in falling object detection in our FADE (the F-measure of FGFA (RAFT) is the second worst).

We think that there are two main reasons for the poor performance of the optical flow based method in FODB task. The first reason is motion blur and occlusion. Appearance of a falling object constantly changes due to the motion blur caused by the fast movement as well as the occlusion caused by trees, neighboring buildings, et al. in the falling process. The second is large displacements. The falling object produces large displacements due to fast movement. With the challenges of motion blur, occlusion, and large displacements, the optical flow is hard to be estimated precisely.

\subsection{Visualization Results}
Visualization results of the representative moving object detection method (MOG2~\cite{zivkovic2004improved}), image object detection method (Faster R-CNN~\cite{ren2015faster}+FPN~\cite{lin2017feature}), video object detection method (MEGA~\cite{chen2020memory}), and our proposed FADE-Net are shown in Figure~\ref{vis}. Many detection results of MOG2 only capture part of the falling object, as MOG2 applies morphological opening in its post-processing, which tends to eliminate small detections.
Thus, some small falling objects on the video frame cannot be detected.
The accuracy of Faster R-CNN + FPN is higher than the other two algorithms. However, this method is also difficult to detect some small falling objects. In addition, the motion blur caused by the fast motion also increases the detection difficulty of the image-based detection model.
The accuracy of MEGA is the worst. This method does not learn the appearance features and motion information of the falling object well during the training process, and the lack of non-maximum suppression processing further reduces its accuracy. 
Our proposed method FADE-Net achieves the best performance in all three video sequences, benefiting from seamlessly integrating generic object detection and motion information in videos. In addition, the designed Moving Attention Module enables the network to assign adaptive weights to different regions based on the motion information, effectively reducing false detections in the image.

\subsection{Effect of incorporating long-term motion information}

Although our FADE-Net primarily focuses on utilizing short-term temporal motion information due to the limited computational capabilities of surveillance camera chips, in this section, we also explore the effectiveness of incorporating long-term motion information into our method.

Specifically, we revise our method to incorporate long-term temporal information to experiment with long-term temporal motion data. The revised method adopts a two-stream architecture: one stream captures short-term motion, while the other models long-term motion across the current frame and the preceding five frames.  Here, we evaluate the revised model on the FADE dataset using a GTX 1080 GPU to simulate a realistic deployment scenario. The experimental results show that the performance of our original model is comparable to that of the revised model (see Table~\ref{tab:longterm}). We attribute this to the high velocity of falling objects, which causes large motion displacements over the long-term window, making it challenging for the model to extract reliable long-term motion cues.
Moreover, we can observe from the experimental results that the revised model incurs a certain loss in inference efficiency compared to the original model (FPS 8.9 vs 15.7), which may limit its applicability in real-world falling object detection scenarios.  Therefore, we focus on utilizing short-term temporal motion information.

\subsection{Effect of introducing domain adaptation techniques into our method}

Since our falling object detection method is intended for real-world deployment and is expected to operate continuously, it may encounter unseen weather scenarios that were not present during training. To address this, we explore the integration of domain adaptation techniques~\cite{chen2022dtt,pan2023cross,guo2022progressive,sun2024d2sl}, commonly used in low-level vision tasks~\cite{zhang2025perform,chen2022dtt,pan2023cross,guo2022progressive,sun2024d2sl} such as image deraining~\cite{chen2022dtt,pan2023cross} and defogging~\cite{guo2022progressive,sun2024d2sl} to improve robustness, into our method and evaluate their effectiveness in this context.

Specifically, we design two variants of our method by integrating it with three different domain adaptation strategies, including an adversarial training-based approach using the Gradient Reversal Layer (GRL)~\cite{ganin2015unsupervised}, as well as a test-time domain adaptation methods, Tent~\cite{wang2020tent}. To better evaluate these variants, we re-partitioned our proposed dataset so that the weather types in the training, validation, and test sets do not overlap. To be specific, the training set contains fair and cloudy conditions, the validation set contains overcast conditions, and the test set contains rainy conditions. Notably, this split is used only for this evaluation.

\begin{table}[t!]
\centering
\vspace{-2mm}
\caption{Effect of incorporating long-term motion information in the proposed FADE-Net.
 The better result is indicated in \textbf{bold}.}
\label{tab:longterm}
\begin{tabular}{cccccc} 
\toprule
Method & F-measure &Precision & Recall & TRO & FPS \\
\midrule
Ours w/o long-term & 72.03& 73.48&\textbf{70.70} &51.75 &\textbf{15.7} \\
\midrule
Ours w/ long-term & \textbf{72.08}& \textbf{73.52}&70.69 &\textbf{51.77} &8.9 \\
\bottomrule
\end{tabular}
\vspace{-2mm}
\end{table}

The experimental results are presented in Table~\ref{tab:domain} above. From the results, we observe that the performance of Ours and Ours w/ GRL is comparable. We believe this is because, although the domain adaptation technique (GRL) encourages the backbone to extract domain-invariant appearance features, in our task where the target objects are extremely small, effectively leveraging motion features is more critical than enhancing appearance features. This observation is further supported by the ablation study, which compares the effectiveness of the motion-focused Moving Attention Module and the appearance-focused Small-Object Mining RPN. In addition, as shown in Table~\ref{tab:domain}, our method benefits from the incorporation of the test-time domain adaptation method Tent. However, since Tent operates during inference, it introduces additional computational overhead and reduces the algorithm's FPS, making it less suitable for real-world deployment. Therefore, considering the overall trade-off between accuracy and inference speed, we adopt ``Ours'' as the final version.

\section{Conclusion}
In this work, we propose a new large-scale video dataset termed FADE for falling object detection around buildings (FODB). It contains 2,611 videos and 245,177 video frames. The videos in FADE include various categories of objects and are captured under diverse scenes, weather conditions, lighting conditions, and video resolutions. Notably, different from the existing MOD datasets, our FADE is the first dataset specialized for FODB. Additionally, we introduce a new baseline method called FADE-Net, which seamlessly integrates motion information capturing and small-sized proposal mining into the detection network. Furthermore, to better evaluate the FODB algorithms, we design an evaluation metric called TRO, which measures the algorithm's ability to locate the beginning and ending times of falling incidents.

We provide a comprehensive benchmark that includes our FADE-Net baseline method, popular MOD methods, generic object detection methods, and video object detection methods. 
Extensive experimental results demonstrate that FODB is a challenging task in the presence of complex backgrounds and motion blur, and validating the effectiveness of our explored baseline method FADE-Net. The FADE dataset, with its diverse videos, will promote the progress of FODB and may also be useful for the investigation of MOD, generic object detection, and video object detection. In future, we will continue to refine and expand the dataset FADE, and explore more advanced FODB methods.

\begin{table}[t!]
\centering
\vspace{-2mm}
\caption{Effect of domain adaptation techniques in FADE-Net. We use a GTX 1080 to evaluate the FPS of the algorithms.
 The better result is indicated in \textbf{bold}.}
\label{tab:domain}
\begin{tabular}{cccccc} 
\toprule
Method & F-measure &Precision & Recall & TRO  &FPS\\
\midrule
Ours  & 72.08& 73.52& 70.69& 51.77  &\textbf{15.7}\\
Ours w/ GRL~\cite{ganin2015unsupervised}  & 72.07& 73.60 &70.65 &51.78 &\textbf{15.7}\\
Ours w/  Tent~\cite{wang2020tent} & \textbf{72.15} & \textbf{73.63} &\textbf{70.73}&\textbf{51.80}  & 10.7 \\
\bottomrule
\end{tabular}
\vspace{-2mm}
\end{table}

\section*{Acknowledgments}
This work was supported by the Natural Science Fund
for Distinguished Young Scholars of Hubei Province under Grant 2022CFA075, the National Natural Science Foundation of China (NSFC) under Grant 62106177, and the Fundamental Research Funds for the Central Universities under Grant 2042023KF0180. The numerical calculation was supported by the super-computing system in the Super-computing Center of Wuhan University.

{
\bibliographystyle{IEEEtran}
\bibliography{bib}
}

\begin{IEEEbiography}[{\includegraphics[width=0.95\textwidth]{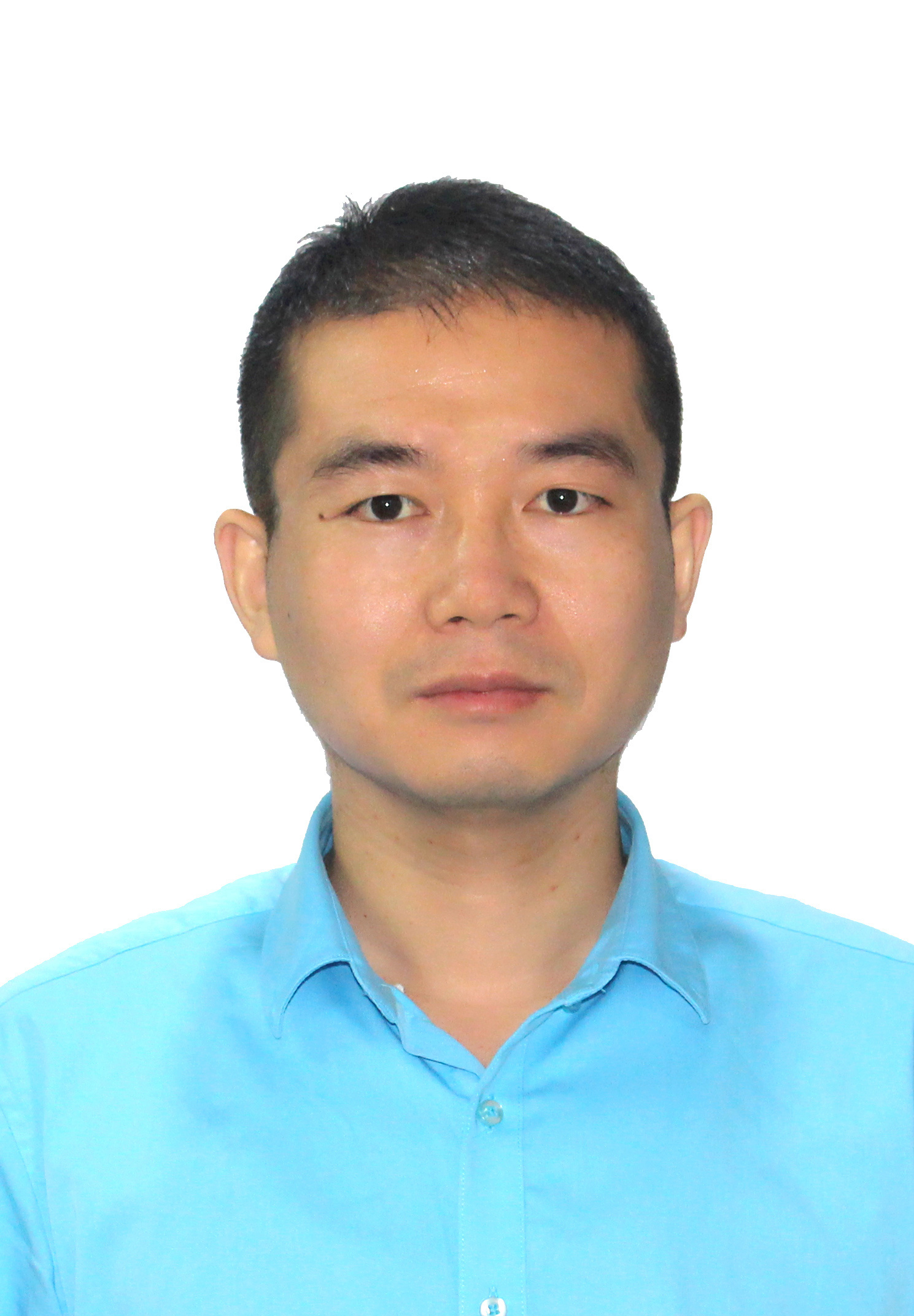}}]
{\textbf{Zhigang Tu}} (Member, IEEE) received the Ph.D.
degree from Wuhan University, China, in 2013, and the Ph.D. degree from Utrecht University, Netherlands, in 2015. From 2015 to 2016, he was a Postdoctoral Researcher with Arizona State University, USA. From 2016 to 2018, he was a Research Fellow with Nanyang Technological University, Singapore. 

He is currently a Professor with Wuhan University. He has
co-/authored more than 70 papers in international SCI-indexed journals and conferences. His current research interests include computer vision, image processing, video analytics, machine learning, motion estimation, human action and gesture recognition, and anomaly event detection. He is the first organizer of the ACCV2020
Workshop on MMHAU, Japan. He received the Best Student Paper Award at the $4^{th}$ Asian Conference on Artificial Intelligence Technology and one of the three best reviewers awards for \textit{IEEE Transactions on Circuits and Systems for Video Technology (IEEE T-CSVT)} in 2022. He is the Area Chair of AAAI2023/2024 and VCIP2022. He is an Associate Editor of the SCI-indexed journal \textit{The Visual Computer} (IF=3.5) and a Guest Editor of \textit{Journal of Visual Communications and Image Representation} (IF=2.6).
\end{IEEEbiography}

\begin{IEEEbiography}[{\includegraphics[width=0.95\textwidth]{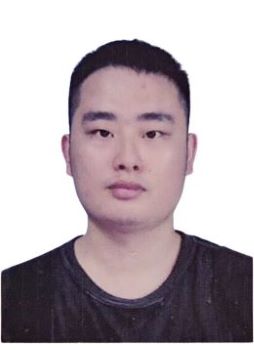}}]
{\textbf{Zhengbo Zhang}} received his Bachelor's and Master's degrees in engineering from Wuhan University. His current research interest is in computer vision.
\end{IEEEbiography}

\begin{IEEEbiography}[{\includegraphics[width=0.95\textwidth]{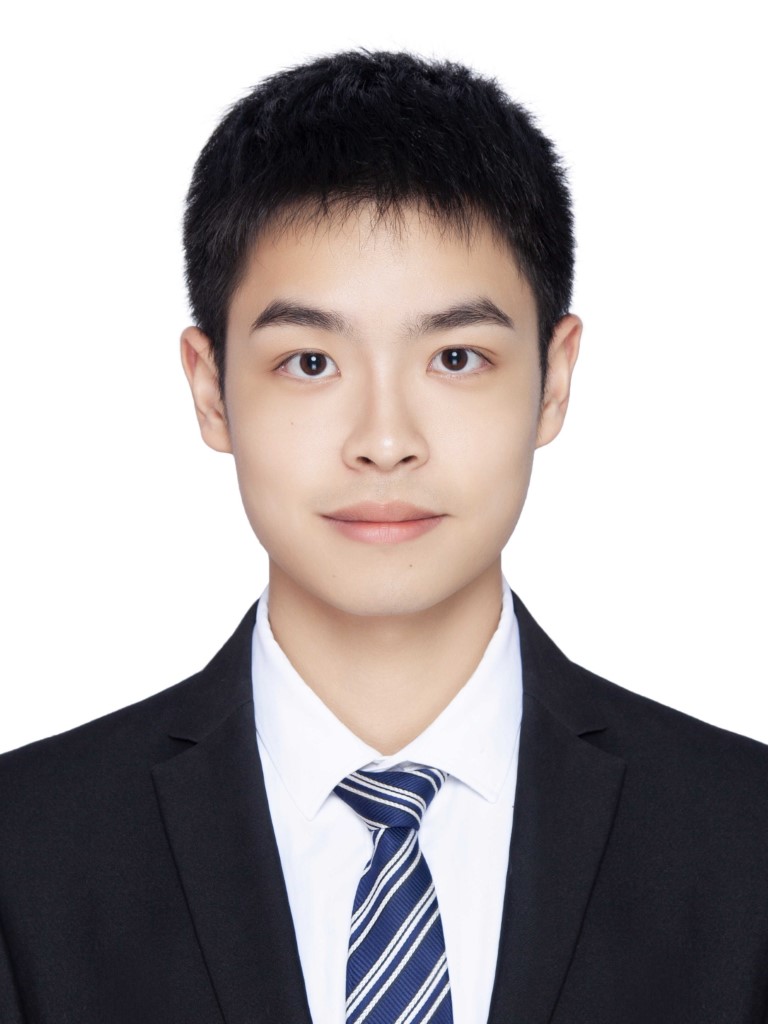}}]
{\textbf{Zitao Gao}} received the Bachelor's degree in engineering from Wuhan University, China, in 2018, and is currently pursuing a Master's degree in the State Key Laboratory of Information Engineering in Surveying, Mapping and Remote Sensing at Wuhan University. He has received honors such as the title of Outstanding Graduate from Wuhan University and the first prize of the Wuhan University Graduate Outstanding Freshman Scholarship. As the team captain, he led the team to win an Honorable Mention in the 2022 CVPR challenge called 'Robustness in Sequential Data'. His current research interests include computer vision, image processing, video analytics, object detection, and human action recognition, and he is actively involved multiple related scientific research projects.
\end{IEEEbiography}

\begin{IEEEbiography}[{\includegraphics[width=0.95\textwidth]{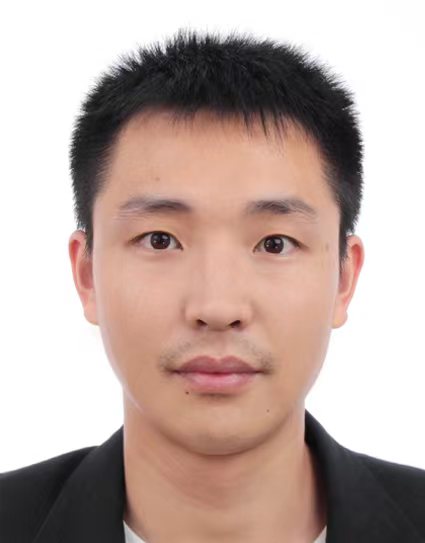}}]
{\textbf{Chunluan Zhou}}
 received the B.Eng. degree from Harbin Institute Technology, China, in 2008, the M.Eng. degree from Zhejiang University, China, in 2011, and the Ph.D. degree from Nanyang Technoglocigal University, Singapore, in 2018. His research interests include object detection, pedestrian detection, visual tracking, pose estimation and tracking, and vision/multi-modal pretraining.
\end{IEEEbiography}

\begin{IEEEbiography}[{\includegraphics[width=0.95\textwidth]{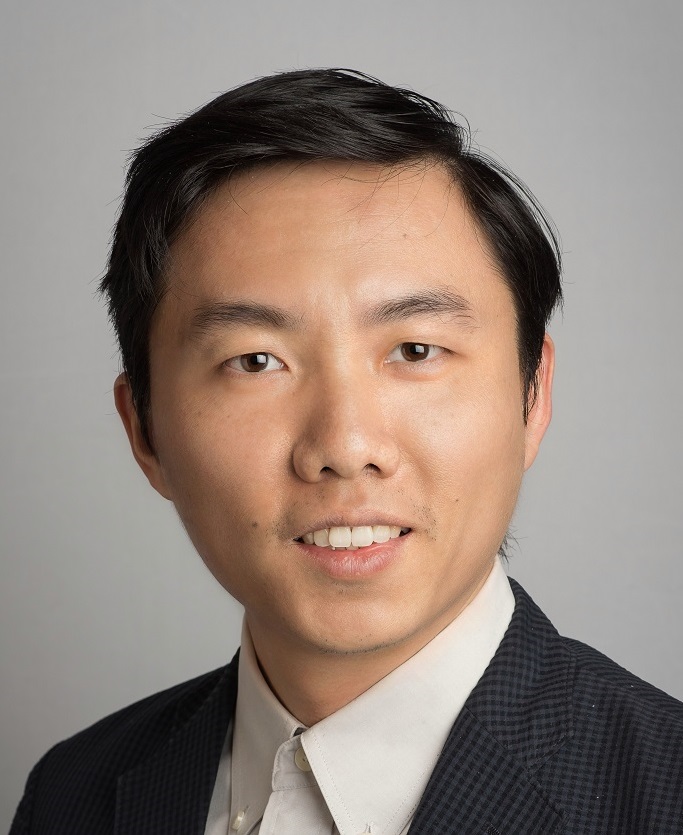}}]
{\textbf{Junsong Yuan}} (Fellow, IEEE) is Professor and Director of Visual Computing Lab at Department of Computer Science and Engineering, State University of New York at Buffalo (UB), USA. Before that he was Associate Professor (2015-2018) and Nanyang Assistant Professor (2009-2015) at Nanyang Technological University (NTU), Singapore. He obtained his Ph.D. from Northwestern University in 2009, M. Eng. from National University of Singapore in 2005, and B. Eng. from Huazhong University of Science Technology in 2002. His research interests include computer vision, pattern recognition, video analytics, large-scale visual search and mining. He received Best Paper Award from IEEE Trans. on Multimedia, Nanyang Assistant Professorship from NTU, and Outstanding EECS Ph.D. Thesis award from Northwestern University. 

He served as Associate Editor of IEEE Trans. on Pattern Analysis and Machine Intelligence (TPAMI), IEEE Trans. on Image Process. (TIP), IEEE Trans. on Circuits and Systems for Video Tech. (TCSVT), and Senior Area Editor of Journal of Visual Communications and Image Representation. He was Program Co-Chair of IEEE Conf. on Multimedia Expo (ICME'18/2022/2024), and Area Chair for CVPR, ICCV, ECCV, and ACM MM. He was elected senator at both NTU and UB. He is a Fellow of IEEE and IAPR.
\end{IEEEbiography}

\begin{IEEEbiography}[{\includegraphics[width=1.1\textwidth]{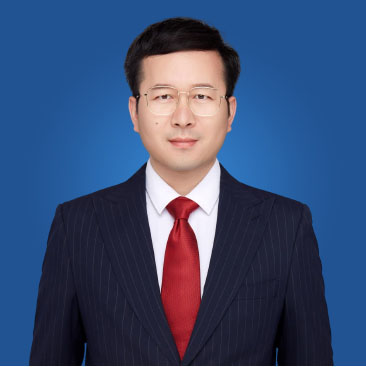}}]
{\textbf{Bo Du}} (Senior Member, IEEE) received the Ph.D. degree in photogrammetry and remote sensing from the State Key Laboratory of Information Engineering in Surveying, Mapping and Remote Sensing, Wuhan University, Wuhan, China, in 2010. He is a Professor with the School of Computer Science, Wuhan University. He has over 80 research articles published in the journals of IEEE Transactions on Pattern Analysis and Machine Intelligence (TPAMI), IEEE Transactions on Image Processing (TIP), IEEE Transactions on Geoscience and Remote Sensing (TGRS), ISPRS Journal of Photogrammetry and Remote Sensing, etc. More than 30 of them are ESI hot articles or highly cited articles. His major research interests include pattern recognition, hyperspectral image processing, machine learning, and signal processing.
\end{IEEEbiography}

\vfill

\end{document}